\documentclass[10pt,twocolumn,letterpaper]{article}
\usepackage{cvpr}
\usepackage{times}
\usepackage{epsfig}
\usepackage{graphicx}
\usepackage{amsmath}
\usepackage{amssymb}
\usepackage{enumitem}
\usepackage{subcaption}
\usepackage{url}
\usepackage{anyfontsize}
\usepackage[usenames,dvipsnames,svgnames,table]{xcolor}
\usepackage[bottom]{footmisc}
\newcommand{\ct}{\textcolor{black}} 
\newcommand{\ctt}{\textcolor{black}} 
\newcommand{\ak}{\textcolor{black}} 
\newcommand{\akk}{\textcolor{black}} 
\newcommand{\akkk}{\textcolor{black}}
\usepackage[pagebackref=true,breaklinks=true,letterpaper=true,colorlinks,bookmarks=false]{hyperref}
\cvprfinalcopy 
\def\cvprPaperID{282}

\ifcvprfinal\fi
\begin{document}
\title{\ct{Seeing Behind the Camera:} Identifying the Authorship of a Photograph}

\author{Christopher Thomas \hspace{2cm} Adriana Kovashka \\
Department of Computer Science\\
University of Pittsburgh\\
{\tt\small \{chris, kovashka\}@cs.pitt.edu}
}

\maketitle
\thispagestyle{empty}

\begin{abstract}
We introduce the novel problem of identifying the photographer behind a photograph. To explore the feasibility of current computer vision techniques to address \ct{this} problem, we created a new dataset of over \ct{180,000} images taken by \ct{41} well-known photographers. Using this dataset, we examined the effectiveness of a variety of features (low and high-level, including CNN features) at identifying the photographer. 
We also trained a new deep convolutional neural network for this task. Our 
results show that high-level features greatly outperform low-level features. 
We provide qualitative results 
using these learned models that give insight into our method's ability
to distinguish between photographers, and allow us to draw interesting
conclusions about what specific photographers shoot. We also demonstrate two applications of our method. 
\end{abstract}

\section{Introduction}
``Motif Number 1'', a simple red fishing shack on the river, is
considered the most frequently painted building in America. Despite
its simplicity, artists' renderings of it vary wildly from
minimalistic paintings of the building focusing on the sunset behind
it to more abstract portrayals of its reflection in the water. This
example demonstrates the great creative license artists have in their
trade, resulting in each artist producing works of art reflective of
their personal style. 
Though the differences may be more subtle, even artists practicing
within the same movement will produce distinct works, owing to
different brush strokes, choice of focus and objects portrayed, use of color, portrayal of
space, and other features emblematic of the individual
artist. While predicting authorship in paintings and classifying painterly style are challenging problems, there have been attempts in
computer vision to automate these tasks \cite{lsc, keren2003recognizing,
johnson2008image, shamir2010impressionism, arora2012towards,
carneiro2012artistic, bar2014classification}. 

While researchers have made progress towards matchings the
human ability to categorize paintings by style and authorship
\cite{lsc, bar2014classification, arora2012towards}, no attempts have been made to recognize the authorship of
\emph{photographs}. This is surprising because the average person is exposed to many
more photographs daily than to paintings.

\begin{figure}[t]
\begin{center}
\begin{subfigure}{.33\columnwidth}
  \centering
  \includegraphics[width=1\columnwidth]{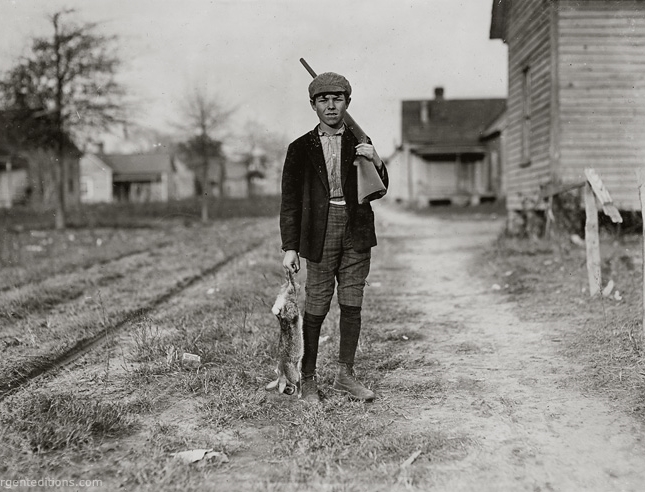}
    \caption{}
  \label{fig:sfig1}
\end{subfigure}%
\begin{subfigure}{.33\columnwidth}
  \centering
  \includegraphics[width=1\columnwidth]{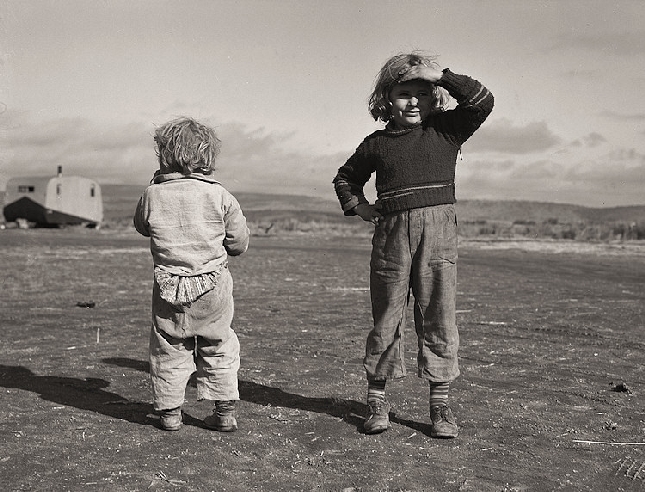}
  \caption{}
  \label{fig:sfig2}
\end{subfigure}
\hspace{-.5em}
\begin{subfigure}{.33\columnwidth}
  \centering
  \includegraphics[width=1\columnwidth]{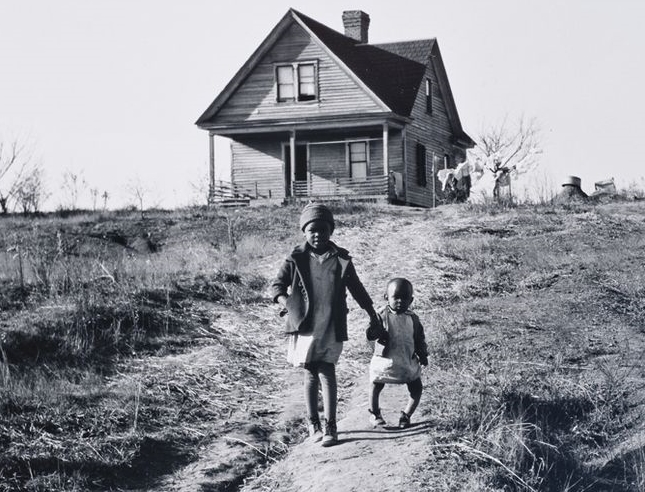}
  \caption{}
  \label{fig:sfig3}
\end{subfigure}
\vspace{-1.5em}
\end{center}
\caption{Three sample photographs from our dataset taken by Hine, Lange, and Wolcott, respectively. Our top-performing feature is able to correctly determine the author of all three photographs, despite the very similar content and appearance of the photos.}
\label{fig:photos}
\vspace{-0.5em}
\end{figure}

Consider again the situation posed in the first paragraph, in which
multiple artists are about to depict the same scene. However this time
instead of painters, imagine that the artists are photographers. In
this case, the stylistic differences previously discussed are not
immediately apparent.
The stylistic cues (such as brush stroke) available for
identifying a particular artist are greatly reduced in the
photographic domain due to the lessened authorial control in that
medium (we do not consider photomontaged or edited images in this
study). 
This makes the problem of identifying the author of a photograph significantly more challenging than that of identifying the author of a painting.

Fig.\ \ref{fig:photos} shows photographs 
taken by Lewis Hine, Dorothea Lange, and Marion Wolcott, three iconic
American photographers.\footnote{Both Lange and Wolcott worked for the Farm
Security Administration (FSA) documenting the hardship of the Great
Depression, while Hine worked to address a number of labor rights
issues.}
All three images depict child poverty and there are no
obvious differences in style,
yet our method is able to correctly predict the author of each. 

\ak{The ability to accurately extract stylistic and authorship information from
artwork computationally enables} a wide array of useful applications in the age of
massive online image databases. 
For example, a user who wants to retrieve more work from a given photographer, 
but does not know his/her name, can speed up the
process by querying with a sample photo and using ``Search
by artist'' functionality that first recognizes the artist.
\akk{Automatic photographer identification can be used to detect unlawful appropriation of others' photographic work, \eg in online portfolios, and could be applied in resolution of intellectual property disputes.
It can also be employed to analyze relations between photographers and discover ``schools of thought'' among them. 
The latter can be used in attributing historical photographs with missing author information.}
\akkk{Finally, understanding a photographer's style might enable the creation of novel photographs in the spirit of a known author.} 

This paper makes several important contributions: 
1) we propose the problem of photographer identification, which no
existing work has explored; 2) due to the lack of a relevant dataset
for this problem, we create a large and diverse dataset which tags each image with its photographer (and \ct{possibly} other metadata); 
3) we investigate a
large number of pre-existing and novel visual features \ct{and their
performance in a comparative experiment in addition to human baselines obtained from a small study; }
4) we provide numerous
qualitative examples and visualizations to illustrate: the features
tested, successes and failures of the method, and interesting
inferences that can be drawn from the learned models; 
5) we apply our method to discover schools of thought between the authors in our dataset; and
6) we show preliminary results on generating novel images that \emph{look like} a given photographer's work.\footnote{\akkk{Automatically creating a novel Rembrandt \emph{painting} \cite{rembrandt} gained media attention in April 2016, five months after we submitted our work.}}

The remainder of this paper is structured as follows. Section
\ref{relwork} presents other research relevant to this problem and
delineates how this paper differs from existing work. Section \ref{dataset} describes the dataset we have
assembled for this project.
Section \ref{features} explains all of the features tested
and how they were \ak{learned}, if applicable. Section \ref{eval} contains
our quantitative evaluation of the different features and
an analysis of the results. Section \ref{qual} provides
qualitative examples, \akk{as well as two applications of our method.}
Section \ref{conclusion} concludes the paper.

\section{Related Work} 
\label{relwork}
The task of automatically determining the author of a particular work
of art has always been of interest to art historians whose job it is
to identify and authenticate newly discovered works of art. 
The problem has been studied by vision researchers, who attempted to
identify Vincent van Gogh forgeries, and to identify distinguishing features
of painters \cite{forgery, farid2009image, johnson2008image, cornelis2009report}. 
While the early application of art analysis was for detecting
forgeries, more recent research has studied how
to categorize paintings by school (\eg, ``Impressionism'' vs
``Secession'') 
\cite{lsc, keren2003recognizing,
johnson2008image, shamir2010impressionism, arora2012towards,
bar2014classification, blessing2010using}. 
\cite{lsc} explored a variety of features and metric learning approaches for computing the similarity between paintings and styles. 
Features based on visual appearance and image transformations have found some success in distinguishing more conspicuous painter and style differences in \cite{blessing2010using, shamir2010impressionism, keren2003recognizing}, all of which explored low level-image features on simple datasets. Recent research has suggested that when coupled with object detection features, the inclusion of low-level features can yield state-of-the-art performance \cite{bar2014classification}.
\cite{arora2012towards} used 
the Classeme \cite{torresani2010efficient} descriptor as their
semantic feature representation.
While it is not obvious that the object detections captured by
Classemes would distinguish painting styles, Classemes outperformed
all of the low-level features.
This indicates that the objects appearing in a painting are also a useful predictor of style. 

Our work also considers authorship identification, but the change of
domain from painting to photography 
\ak{poses novel challenges that demand a different solution than that which was applied for painter identification.} 
The distinguishing features of
  painter styles (paint type, smooth or hard brush, etc.) are
inapplicable to the photography domain.
Because the photographer lacks
the imaginative canvas of the painter, variations in photographic
style are much more subtle. Complicating matters further, many of the
photographers in our dataset are from roughly the same time period,
some even working for the same government agencies with the same
stated job purpose. 
Thus, photographs taken by the subjects tend to be very similar in
appearance and content, making distinguishing them particularly
challenging, even for humans.

\begin{table*}[t]
\vspace{-.2em}
\begin{center}
\resizebox{\textwidth}{!}{
\begin{tabular}{|c|c|c|c|c|c|c|c|c|c|c|c|cc|}
\hline
Adams     & 245   & Brumfield & 1138 & Capa     & 2389 & Bresson    & 4693 & Cunningham & 406   & Curtis          & 1069  & \multicolumn{1}{c|}{Delano}    & 14484 \\ \hline
Duryea    & 152   & Erwitt    & 5173 & Fenton   & 262  & Gall       & 656  & Genthe     & 4140  & Glinn           & 4529  & \multicolumn{1}{c|}{Gottscho}  & 4009  \\ \hline
Grabill   & 189   & Griffiths & 2000 & Halsman  & 1310 & Hartmann   & 2784 & Highsmith  & 28475 & Hine            & 5116  & \multicolumn{1}{c|}{Horydczak} & 14317 \\ \hline
Hurley    & 126   & Jackson   & 881  & Johnston & 6962 & Kandell    & 311  & Korab      & 764   & Lange           & 3913  & \multicolumn{1}{c|}{List}      & 2278  \\ \hline
McCurry   & 6705  & Meiselas  & 3051 & Mydans   & 2461 & O'Sullivan & 573  & Parr       & 20635 & Prokudin-Gorsky & 2605  & \multicolumn{1}{c|}{Rodger}    & 1204  \\ \hline
Rothstein & 12517 & Seymour   & 1543 & Stock    & 3416 & Sweet      & 909  & Van Vechten    & 1385  & Wolcott         & 12173 &                                &       \\ \hline
\end{tabular}
}
\end{center}
\vspace{-1.5em}
\caption{Listing of all photographers and the number of photos by each
  in our dataset.}
\label{tab:dataset}
\vspace{-0.5em}
\end{table*}

There has been work in computer vision that studies aesthetics in photography \cite{marchesotti2011assessing, murray2012ava, dhar2011high}.
Some work also studies style in architecture \cite{doersch2012makes, lee2015linking}, vehicles \cite{lee2013style}, or yearbook phootgraphs \cite{ginosar2015century}.
However, all of these differ from our goal of \emph{identifying authorship} in photography.
Most related to our work is the study of visual style in
  photographs, 
  conducted by \cite{ris}. 
  Karayev \etal conducted a \ak{broad} study
  on both paintings and photographs. The 20 style classes and 25 art
  genres considered in their study are coarse (HDR, Noir, Minimal,
  Long Exposure, etc.) and much easier to distinguish than the
  photographs in our dataset, many of which are of the same types of
  content and have very similar visual appearance. 
  While \cite{ris} studied
  style in the context of photographs and paintings, we explore the
  novel problem of \emph{photographer identification}. We find it unusual
  that this problem has remained unexplored for so long, given 
  \ak{that photographs are more abundant than paintings, and there has been work in computer vision to analyze paintings}. 
  Given the lower level of authorial control
  that the photographer possesses compared to the painter, we believe
  that the photographer classification task is more challenging, in
  that it often requires attention to subtler cues than brush stroke, for example. Besides our experimental analysis of this new
  problem, we also contribute the first
  large dataset of well-known photographers and
  their work.
  
\akkk{In Sec. \ref{sec:gen}, we propose a method for generating a new photograph in the style of an author. This problem is distinct from style transfer \cite{bae2006two, bychkovsky2011learning, aubry2014fast} which adjusts the tone or color of a photograph. Using \cite{aubry2014fast} on our generated photographs did not produce a visible improvement in their quality.}

\section{Dataset} \label{dataset}

A significant contribution of this paper is our photographer
dataset.\footnote{It can be downloaded at \url{http://www.cs.pitt.edu/~chris/photographer}.} It consists of 41 well known photographers and
contains 181,948 images of varying resolutions. We searched Google for ``famous photographers'' and used the list while
also choosing authors with large, curated collections available online. Table
\ref{tab:dataset} contains a listing of each photographer and their
associated number of images in our dataset. The timescale of the
photos spans from the early days of photography to the present day. As
such, some photos have been developed from film and some are
digital. \ct{Many} of the images were harvested using a web spider with
permission from the Library of Congress's photo archives \ct{and the} National Library of Australia's
digital collection's website. \ct{The rest were harvested from the Magnum Photography online catalog, or from independent photographers' online collections.} Each photo in the dataset is annotated
with the ID of the author, the URL from which it was obtained, \ct{and possibly other meta-data, including: the title of the photo, a
summary of the photo, and the subject of the photo (if known).} The title, summary, and subject of the
photograph were provided by either the curators of the collection or
by the photographer. Unlike other datasets obtained through
web image search which may contain some incorrectly labeled
images, our dataset has been painstakingly assembled, authenticated,
and described by the works' curators. This rigorous process ensures
that the dataset and its associated annotations are of the highest
quality. 

\section{Features} \label{features}
Identification of the correct photographer is a complex problem
and relies on multiple factors. Thus, we explore
a broad space of features (both low and high-level). The
term ``low-level'' means that each dimension of the feature
vector has no inherent ``meaning.'' High-level features have articulatable semantic meaning (i.e.\ the presence of an object in the image). 
We also train a deep convolutional neural network from scratch in order to learn custom features specific to this problem domain.

\begin{table*}[t]
\small
\begin{center}
\vspace{-0.2em}
\resizebox{\textwidth}{!}{
\begin{tabular}{|c|c|c|c|c|c|c|c|c|c|c|c|c|c|c|c|c|}
\hline
\multicolumn{3}{|c|}{\textbf{Low}} & \multicolumn{14}{c|}{\textbf{High}}
\\ \hline
\multicolumn{3}{|c|}{}                                   &             & \multicolumn{4}{c|}{\textbf{CaffeNet}} & \multicolumn{4}{c|}{\textbf{Hybrid-CNN}} & \multicolumn{5}{c|}{\textbf{PhotographerNET}} \\ \hline
Color            & GIST            & SURF-BOW              & Object Bank & Pool5     & FC6     & FC7     & FC8    & Pool5     & FC6      & FC7     & FC8     & Pool5    & FC6    & FC7     & FC8    & TOP    \\ \hline
0.31             & 0.33            & 0.37                  & 0.59        & \emph{0.73}      & 0.7     & 0.69    & 0.6    & \emph{\textbf{0.74}}      & 0.73     & 0.71    & 0.61    & 0.25        & 0.25      & \emph{0.63}    & 0.47   & 0.14   \\ \hline
\end{tabular}
}
\end{center}
\vspace{-1.5em}
\caption{Our experimental results. The F-measure
  of each feature is reported. \ak{The best feature overall is in \textbf{bold}, and the best one per CNN in \emph{italics}.}
Note that high-level features
  greatly outperform low-level ones. Chance performance is \ak{0.024}.} 
\vspace{-0.5em}
\label{table:results}
\end{table*}

\vspace{-.5em}
\noindent \textbf{\\ Low-Level Features} 
\begin{itemize}[leftmargin=*]
\vspace{-.5em}
\item \textbf{L*a*b* Color Histogram:} To capture color differences among the photographers, we use a 30-dimensional binning of the L*a*b* color space. Color has been shown useful for dating historical photographs \cite{palermo2012dating}.
\vspace{-.5em}
\item \textbf{GIST:} GIST \cite{oliva2006building} features have been shown to perform well at scene classification and have been tested by many of the prior studies in style and artist identification \cite{ris, lsc}. All images are resized to 256 by 256 pixels prior to having their GIST features extracted.
\vspace{-.5em}
\item \ak{\textbf{SURF:}} Speeded-up Robust Features (SURF) \cite{bay2008speeded}
  is a classic local feature used to find patterns in images and has been used as a baseline for artist and style identification \cite{bar2014classification, blessing2010using, arora2012towards}. 
We use $k$-means clustering to obtain a vocabulary of 500 visual words and apply a standard bag-of-words approach using normalized histograms.
\end{itemize}
\vspace{-1em}
\noindent \textbf{\\ High-Level Features}
\begin{itemize}[leftmargin=*]
\vspace{-.5em}
\item \textbf{Object Bank: } The Object Bank \cite{li2010object} descriptor captures the location of numerous object detector responses. We believe that the spatial relationships between objects may carry some semantic meaning useful for our task.
\vspace{-.5em}
\item \textbf{Deep Convolutional Networks:} 
\begin{itemize}[leftmargin=*]
\vspace{-.5em}
\item \textbf{CaffeNet: } This pre-trained CNN \cite{jia2014caffe} is a clone of the winner of the ILSVRC2012 challenge \cite{krizhevsky2012imagenet}. The network was trained on approximately 1.3M images to classify images into 1000 different object categories.
\item \textbf{Hybrid-CNN: } This network 
has recently achieved state-of-the-art performance on scene recognition benchmarks \cite{zhou2014learning}. It was trained to recognize 1183 scene and object categories on roughly 3.6M images.
\item \textbf{PhotographerNET: } We trained a CNN with the same architecture as the previous networks to identify the author of photographs from our dataset. The network was trained for 500,000 iterations on 4 Nvidia K80 GPUs on our training set and validated on a set \ak{disjoint from our training and test sets}.
\end{itemize}

\noindent 
To disambiguate layer names, we prefix them with a C, H, or P depending on whether the feature came from CaffeNet, Hybrid-CNN, or PhotographerNET, respectively.
For all networks, 
we \akkk{extract features from the Pool5, FC6, FC7 and FC8 layers, and show the result of using those features during SVM training} in Table \ref{table:results}. 
The score in the TOP column for PhotographerNET is produced by classifying each test image as the author who corresponds to the dimension with the maximum response value in PhotographerNET's output (FC8).
\end{itemize}

\section{Experimental Evaluation} \label{eval}
To tested the effectiveness of the aforementioned features on the
photographer classification task, using our new photographer dataset.  
We randomly divided our dataset into a training set (90\%) and test set (10\%). Because a validation set is useful when training a CNN to determine when learning has peaked, we created a validation set by randomly sampling 10\% of the images from the training set and excluding them \ak{from the training set for our CNN only}. 
\ak{The training of our PhotographerNET was terminated when performance started dropping on the validation set.} 

For every feature  in Table \ref{table:results} (except TOP which assigns the max output in FC8 as the photographer label) we train a one-vs-all multiclass SVM using the framework provided by \cite{fan2008liblinear}. All SVMs use linear kernels. 

Table \ref{table:results} presents the results of our experiments. We
report the F-measure for each of the features tested. We observe that
the deep features significantly outperform all low-level standard vision features, concordant with the findings of \cite{ris, bar2014classification, lsc}. Additionally, we observe that Hybrid-CNN features
outperform CaffeNet by a small margin on all features tested. This suggests that while
objects are clearly useful \ct{for photographer identification} given the impressive performance of
CaffeNet, the added scene information of Hybrid-CNN provides useful
cues beyond those available in the purely object-oriented
model. 
We observe that Pool5 is the best feature within both CaffeNet and Hybrid-CNN. 
\akkk{Since Pool5 roughly corresponds to parts of objects \cite{zeiler2014visualizing, wei2015understanding, huang2015salicon}, we can conclude that}
seeing the \emph{parts} of objects, not the \emph{full} objects, is most discriminative for identifying photographers. This is intuitive because an artistic photograph contains many objects, so some of them may not be fully visible.

\ct{The Object Bank feature achieves nearly the same performance as C-FC8 and H-FC8, the network layers with explicit semantic meaning. All} three of these features encapsulate 
object information, 
though Object Bank detects significantly
fewer classes (177) than Hybrid-CNN (978) \ct{or CaffeNet (1000).} Despite detecting fewer
categories, Object Bank encodes more fine-grained spatial information about where the objects detected were
located in the image, compared to \ct{H-FC8 and C-FC8.} This finer-grained information could be giving
it a slight advantage over \ct{these CNN object detectors, despite its fewer categories.}

One surprising result from our experiment is that PhotographerNET does not surpass either CaffeNet or Hybrid-CNN, which were trained for object and scene detection on different datasets.\footnote{We also tried fine-tuning the last three layers of CaffeNet and Hybrid-CNN with our photographer data, but we did not obtain an increase in performance.} 
PhotographerNET's top-performing feature (FC7) outperforms the deepest (FC8) layers in both CaffeNet and Hybrid-CNN, which correspond to object and scene classification, respectively.
However, P-FC7 performs worse than their shallower layers, especially H-Pool5.
Layers of the network shallower than P-FC7, such as P-FC6 and P-Pool5, demonstrate a sharp decrease in performance (a trend opposite to what we see for CaffeNet and Hybrid-CNN), suggesting that PhotographerNET has learned different and less predictive intermediate feature extractors for these layers than CaffeNet or Hybrid-CNN.
\akkk{Attributing a photograph to the author with highest P-FC8 response (TOP) is even weaker because unlike the P-FC8 method, it does not make use of an SVM.}
It may be that the task PhotographerNET is trying to learn is too high-level and challenging. Because PhotographerNET is learning a task even more high-level than object classification and we observe that the full-object-representation is not very useful for this task, one can conclude that for photographer identification, there is a mismatch between the high-level nature of the task, and the level of representation that is useful.

\begin{figure}[t]
\begin{center}
\begin{subfigure}{1\columnwidth}
\centering
\includegraphics[width=1\columnwidth]{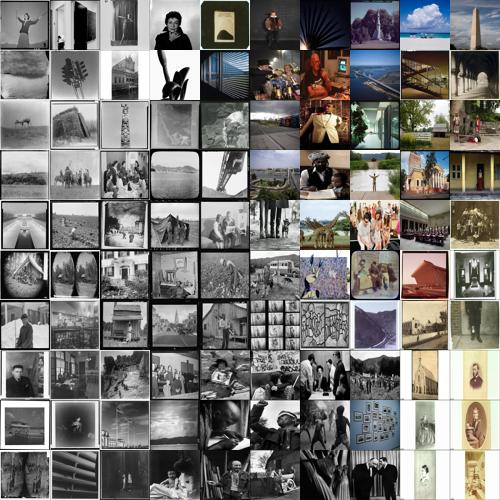}
\caption{P-FC7 t-SNE embeddings.}
\end{subfigure}%
\end{center}
\vspace{-1.5em}
\begin{center}
\begin{subfigure}{1\columnwidth}
\centering
\includegraphics[width=1\columnwidth]{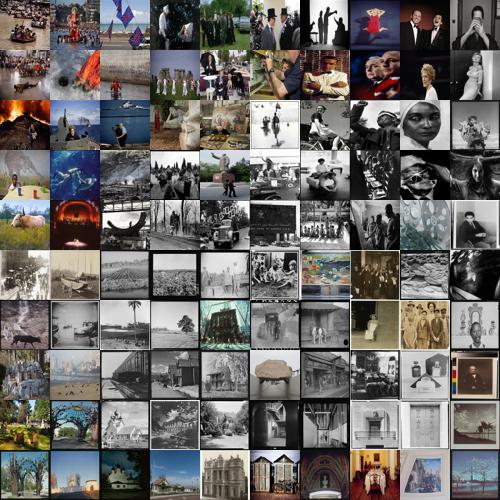}
\caption{H-Pool5 t-SNE embeddings.}
\end{subfigure}%
\end{center}
\vspace{-1.5em}
\caption{t-SNE embeddings for two deep features. We observe that PhotographerNET relies more heavily on lower-level cues (like color) than higher-level semantic details.}
\label{fig:tsne}
\vspace{-1em}
\end{figure}

\akkk{In Fig. \ref{fig:tsne}, we provide a visualization that might explain the relative performance of our top-performing PhotographerNET feature (P-FC7) and the best feature overall (H-Pool5).}
We compute the t-distributed stochastic neighborhood embeddings \cite{van2008visualizing} for P-FC7 and H-Pool5. We use the embeddings to project each feature into 2-D space. We then plot the embedded features by representing them with their corresponding photographs. 

We observe that H-Pool5 divides the image space in semantically meaningful ways. For example, we see that photos containing people are grouped mainly at the top right, while buildings and outdoor scenes are at the bottom. We notice H-Pool5's groupings are agnostic to color or border differences. 
In contrast, PhotographerNET's P-FC7 divides the image space along the diagonal into black and white vs.\ color regions. It is hard to identify semantic groups based on the image's content. However, we can see that images that ``look alike'' by having similar borders or similar colors are closer to each other in the projection. This indicates that PhotographerNET learned to use lower-level features to perform photographer classification, whereas Hybrid-CNN learned higher-level semantic features for object/scene recognition. One possible explanation for this is that because the photos within each class (photographer) of our dataset are so visually diverse, the network is unable to learn semantic features for objects which do not occur frequently enough. In contrast, networks trained explicitly for object recognition only see images of that object in each class, enabling them to more easily learn object representations. Interestingly, these semantic features learned on a different problem outperform the features learned on our photographer identification problem.

\ct{To establish a human baseline for the task of photographer identification, we performed two small pilot experiments. We created a website where participants could view \ak{50 randomly chosen images training images for} each photographer. The participants were asked to review these and were allowed to take notes. Next, they were asked to classify 30 photos chosen at random from a special balanced test set. Participants were allowed to keep open the page containing the images for each photographer during the test phase of the experiment.} \ct{In our first experiment, one participant studied \ak{and classified} images for all 41 photographers and obtained an F1-score of 0.47. In a second \ak{study}, a different participant performed the same task but was only asked to study and classify the ten photographers with the most data, and obtained an F1-score of 0.67. 
Our top-performing feature's performance in Table \ref{table:results} (on all 41 photographers) surpasses both human F1-scores even on the smaller task of ten photographers, demonstrating the difficulty of the photographer identification problem on our challenging dataset.}

\begin{figure}[]
\begin{center}
\begin{subfigure}{.33\columnwidth}
\centering
\includegraphics[width=1\columnwidth]{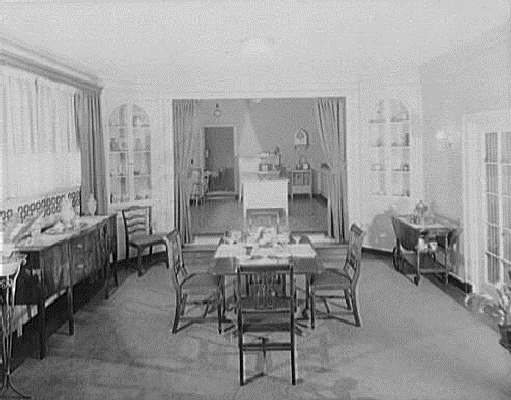}
\caption{Horydczak}
\label{fig:confusedimages:d}
\end{subfigure}%
\begin{subfigure}{.33\columnwidth}
\centering
\includegraphics[width=1\columnwidth]{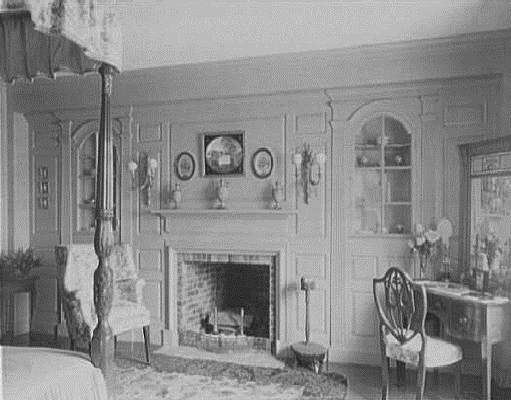}
\caption{Gottscho-SURF}
\label{fig:confusedimages:e}
\end{subfigure}
\hspace{-.5em}
\begin{subfigure}{.33\columnwidth}
\centering
\includegraphics[width=1\columnwidth]{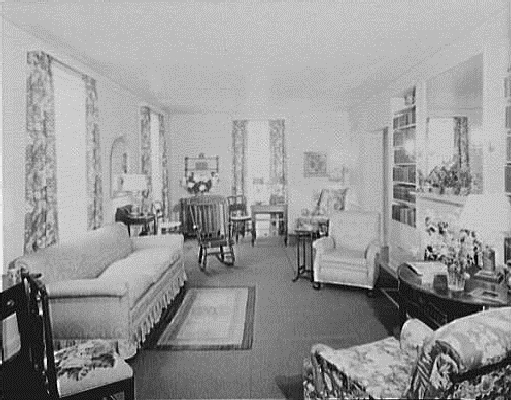}
\caption{Horydczak-SURF}
\label{fig:confusedimages:f}
\end{subfigure}

\begin{subfigure}{.33\columnwidth}
\centering
\includegraphics[width=1\columnwidth]{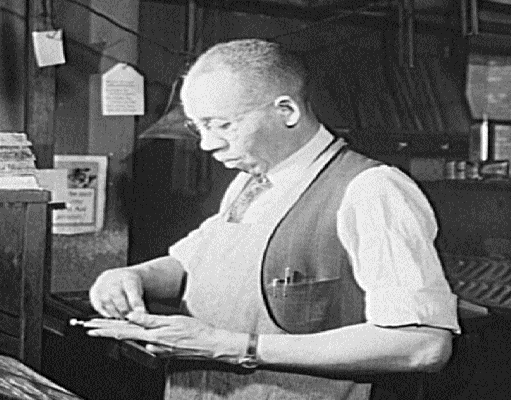}
\caption{Delano}
\label{fig:confusedimages:j}
\end{subfigure}%
\begin{subfigure}{.33\columnwidth}
\centering
\includegraphics[width=1\columnwidth]{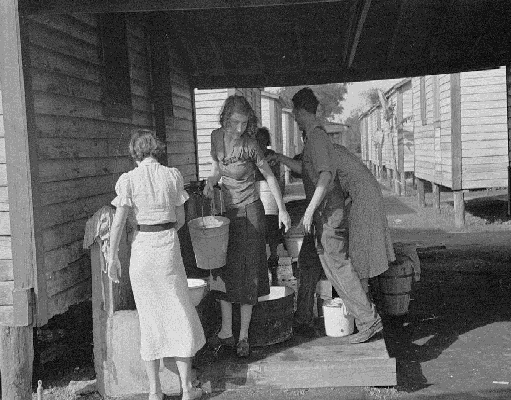}
\caption{Roths.-C-Pool5}
\label{fig:confusedimages:k}
\end{subfigure}%
\begin{subfigure}{.33\columnwidth}
\centering
\includegraphics[width=1\columnwidth]{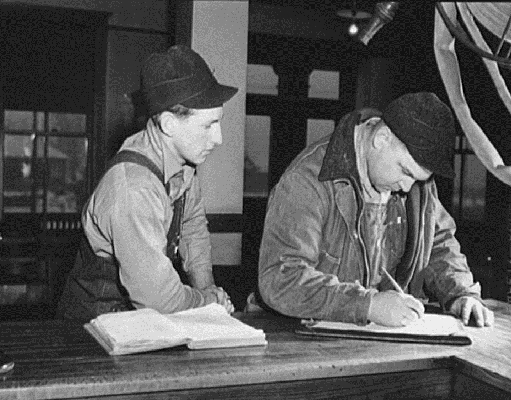}
\caption{Delano-C-Pool5}
\label{fig:confusedimages:l}
\end{subfigure}%

\begin{subfigure}{.33\columnwidth}
\centering
\includegraphics[width=1\columnwidth]{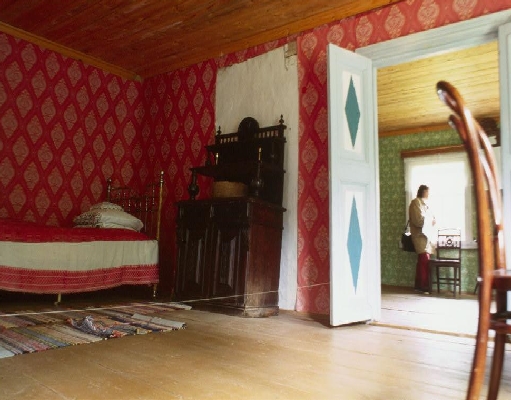}
\caption{Brumfield}
\label{fig:confusedimages:m}
\end{subfigure}%
\begin{subfigure}{.33\columnwidth}
\centering
\includegraphics[width=1\columnwidth]{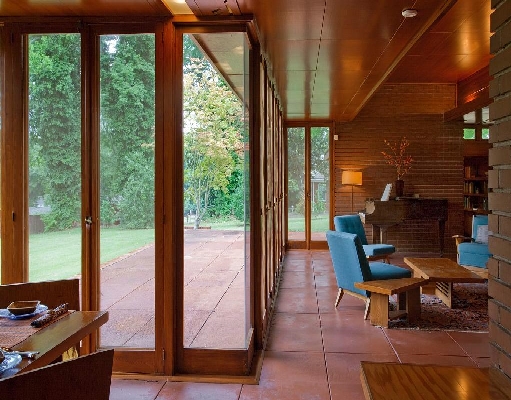}
\caption{High.-H-Pool5}
\label{fig:confusedimages:n}
\end{subfigure}%
\begin{subfigure}{.33\columnwidth}
\centering
\includegraphics[width=1\columnwidth]{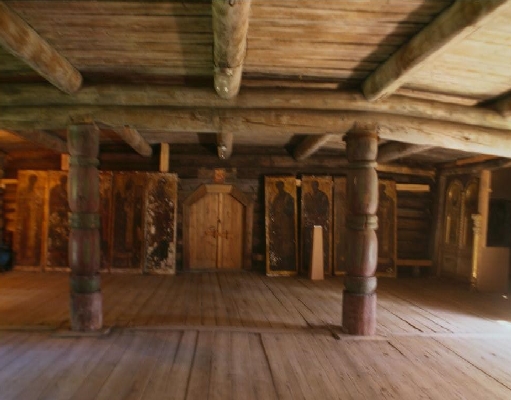}
\caption{Brum.-H-Pool5}
\label{fig:confusedimages:o}
\end{subfigure}%
\end{center}
\vspace{-1.5em}
\caption{Confused images. The first column shows the test image, the second shows the closest image in the predicted class, and the third shows the closest image from the correct class. \ct{Can you tell which one doesn't belong?}}
\label{fig:confusedimages}
\vspace{-1.5em}
\end{figure}

Finally, 
to demonstrate the difficulty of the photographer classification problem and to explore the types of errors different features tend to make, we present several examples of misclassifications in Fig.\  \ref{fig:confusedimages}. 
Test images are shown on the left. Using the SVM weights to weigh
image descriptors, we find the training image (1) from the incorrectly
predicted class (shown in the middle)
and (2) from the correct class (shown on the right), with minimum distance to the
test image. 
The first row (Fig.\ \ref{fig:confusedimages:d}-\ref{fig:confusedimages:f}) depicts confusion using SURF features. All three rooms have visually similar decor and furniture, offering some explanation to
Fig.\ \ref{fig:confusedimages:d}'s misclassification as a Gottscho image. 
The second row (Fig.\ \ref{fig:confusedimages:j}-\ref{fig:confusedimages:l}) shows a misclassification by CaffeNet. Even though all three scenes contain \akk{people at work,} CaffeNet lacks the ability to differentiate between the scene types (indoor vs.\ outdoor and place of business vs.\ house). In contrast, Hybrid-CNN was explicitly trained to differentiate these types of scenes. The final row shows the type of misclassification made by our top-performing feature, H-Pool5. Hybrid-CNN has confused the indoor scene in Fig.\ \ref{fig:confusedimages:m} as a Highsmith. However, we can see that Highsmith took a similar indoor scene containing similar home furnishings (Fig.\ \ref{fig:confusedimages:n}). 
These examples illustrate a few of the many confounding factors 
\akkk{which make photographer identification challenging.}

\section{Qualitative Results} \label{qual}
The experimental results presented in the previous section indicate
that classifiers can exploit semantic information in photographs to
differentiate between photographers at a much higher fidelity than low-level features. At this point, the question becomes not \emph{if}
computer vision techniques can perform photographer classification
relatively reliably but \emph{how} they are doing it. What did the
classifiers learn? In this section, we present qualitative results
which attempt to answer this question and enable us
to draw interesting insights about the photographers and their
subjects.
 
\subsection{Photographers and objects}

\begin{figure}[t]
\begin{center}
\includegraphics[width=0.98\columnwidth]{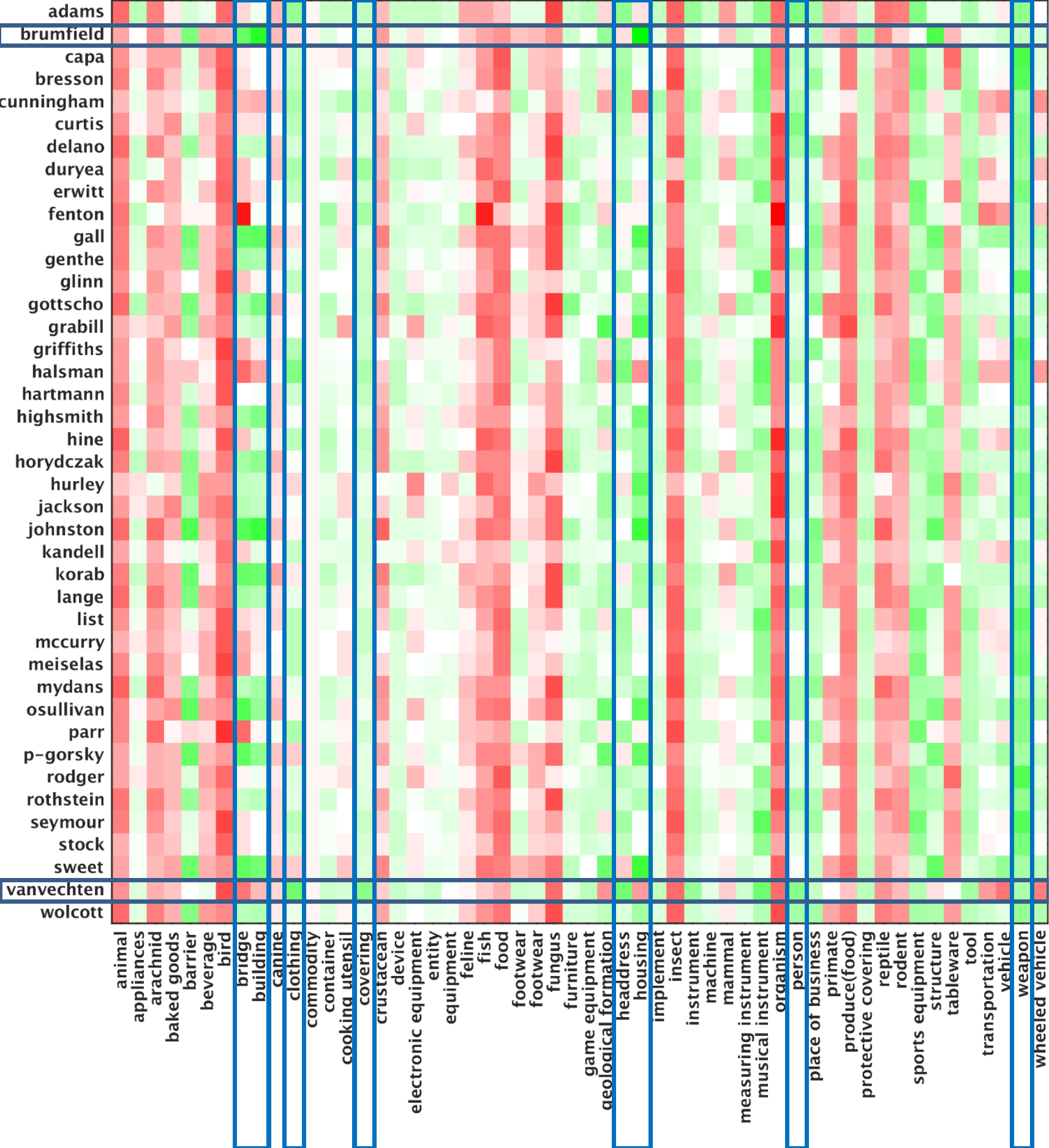}
\end{center}
\vspace{-1.5em}
\caption{Average C-FC8 collapsed by WordNet. \ak{Please zoom in or view the supplementary file for a larger image.}}
\label{fig:photographerobjects}
\vspace{-1em}
\end{figure}

Our first set of qualitative experiments explores the relationship of
each photographer to the objects which they photograph and which
differentiate them. Each dimension of the 1000-dimensional C-FC8 vector produced
by CaffeNet represents a probability that its associated ImageNet
synset is the class portrayed by the image. 
While C-FC8 does not achieve the highest F-measure, it has a clear semantic mapping to
ImageNet synsets and thus can be more easily used to
reason about what the classifiers have learned. 
Because the C-FC8 vector is
high-dimensional, we ``collapse'' the vector for purposes of human
consideration. To do this, we map each ImageNet synset to its
associated WordNet synset and then move up the WordNet hierarchy until the first of a number of manually chosen synsets\footnote{These synsets were manually chosen to form a natural human-like grouping of the 1000 object categories. 
Because the manually chosen synsets are on multiple levels of the WordNet hierarchy,
synsets are assigned to their deepest parent.} 
are encountered, which becomes the dimension's new label.
This reduces C-FC8 to \ct{54} coarse categories
by averaging all dimensions with the same \akk{coarse} label. 
In Fig.\ \ref{fig:photographerobjects}, we show the average response values
for these \ct{54} coarse object categories for each photographer. 
Green indicates positive values and red indicates negative values. Darker shades of each color are more extreme.

\begin{figure}[t]
\begin{center}
\includegraphics[width=0.98\columnwidth]{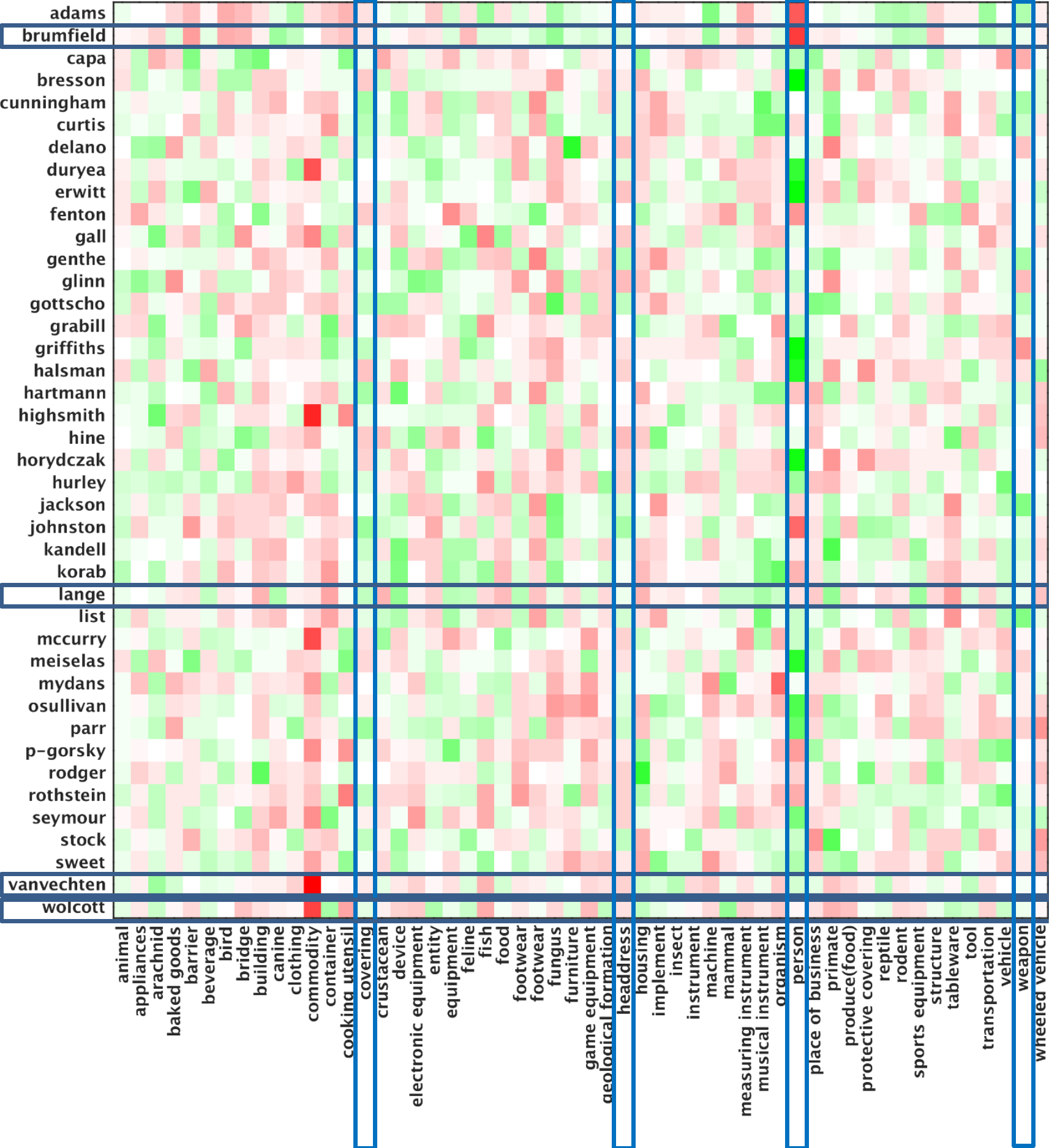}
\end{center}
\vspace{-1.5em}
\caption{C-FC8 SVM weights collapsed by WordNet. \ak{Please zoom in or view supplementary for a larger image.}}
\label{fig:svmobjects}
\vspace{-1.5em}
\end{figure}

We apply the same technique to collapse the learned SVM
weights. During training, each one-vs-all linear SVM learns a weight
for each of the 1000 C-FC8 feature dimensions. Large positive or negative values
indicate a feature that is highly predictive. Unlike the
previous technique which simply shows the average object distribution per photographer, 
using the learned weights allows us to see what categories
specifically \emph{distinguish} a photographer from others.
We show the result in Fig.\ \ref{fig:svmobjects}.

Finally, while information about the \ct{54} \emph{types} of objects photographed by each author is useful, finer-grained detail
is also available. We list the
top 10 individual categories with highest H-FC8 weights (which captures both objects
and scenes). To do this, we extract and average the H-FC8 vector for
all images in the dataset for each photographer. 
We list the top 10 most represented categories for a select
group of photographers in Table \ref{table:topten}, and include example
photographs by each photographer.


We make the following observations about the photographers' style from Figs.\  \ref{fig:photographerobjects} and  \ref{fig:svmobjects} and Table \ref{table:topten}.
From Fig.\ \ref{fig:photographerobjects}, we conclude that Brumfield shoots significantly fewer people than most photographers. 
Instead, Brumfield shoots many ``buildings'' and ``housing.'' Peering deeper, Brumfield's top ten categories in Table \ref{table:topten} reveal that he frequently shot architecture (such as mosques and stupas). In fact, Brumfield is an architectural photographer, particularly of Russian architecture. 
In contrast, Van Vechten has high response values for categories such as ``clothing'', ``covering'', ``headdress'' and ``person''.  Van Vechten's photographs are almost exclusively portraits of people, so we observe a positive SVM weight for ``person'' in Fig.\ \ref{fig:svmobjects}.

Comparing 
 Figs.\ \ref{fig:photographerobjects} and \ref{fig:svmobjects}, 
we see that there is not a clear correlation between object frequency and the object's
  SVM weight. \ct{For instance, the ``weapon'' category is frequently represented \ct{given} Fig.\ \ref{fig:photographerobjects}, yet is only predictive of a
  few photographers (Fig.\ \ref{fig:svmobjects}).} 
The ``person'' category in Fig.\ \ref{fig:svmobjects} has high magnitude weights for many photographers, indicating its utility as a class predictor. 
Note that the set of objects distinctive for a photographer does not fully depend on
  the photographer's environment. For example, Lange and Wolcott both
  worked for the FSA, yet there are notable differences between their SVM weights in Fig.\  \ref{fig:svmobjects}. 


\begin{table*}[t]
\begin{center}
\resizebox{\textwidth}{!}{
\begin{tabular}{|c|c|c|c|c|c|c|c|c|c|c|}
\hline
{\bf Adams}       & hospital room & hospital          & office            & mil. uniform  & bow tie      & lab coat     & music studio & art studio    & barbershop    & art gallery   \\ \hline
{\bf Brumfield}   & dome          & mosque            & bell cote         & castle        & picket fence & stupa        & tile roof    & vault         & pedestal      & obelisk       \\ \hline
{\bf Delano}      & hospital      & construction site & railroad track    & slum          & stretcher    & barbershop   & mil. uniform & train station & television    & crutch        \\ \hline
{\bf Hine}        & mil. uniform  & pickelhaube       & prison            & museum        & slum         & barbershop   & milk can     & rifle         & accordion     & crutch        \\ \hline
{\bf Kandell}     & flute         & marimba           & stretcher         & assault rifle & oboe         & rifle        & panpipe      & cornet        & mil. uniform  & sax           \\ \hline
{\bf Lange}       & shed          & railroad track    & construction site & slum          & yard         & cemetery     & hospital     & schoolhouse   & train railway & train station \\ \hline
{\bf Van Vechten} & bow tie       & suit              & sweatshirt        & harmonica     & neck brace   & mil. uniform & cloak        & trench coat   & oboe          & gasmask       \\ \hline
\end{tabular}
}
\end{center}
\vspace{-0.1em}
\centering
\begin{subfigure}{.13\textwidth}
\centering
\includegraphics[width=1\columnwidth]{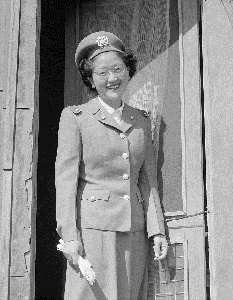}
\caption*{Adams}
\end{subfigure}
\hspace{-.3em}
\begin{subfigure}{.13\textwidth}
\centering
\includegraphics[width=1\columnwidth]{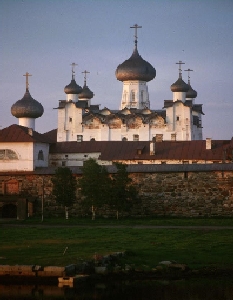}
\caption*{Brumfield}
\end{subfigure}
\hspace{-.3em}
\begin{subfigure}{.13\textwidth}
\centering
\includegraphics[width=1\columnwidth]{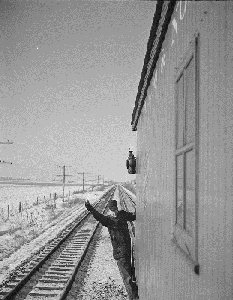}
\caption*{Delano}
\end{subfigure}
\hspace{-.3em}
\begin{subfigure}{.13\textwidth}
\centering
\includegraphics[width=1\columnwidth]{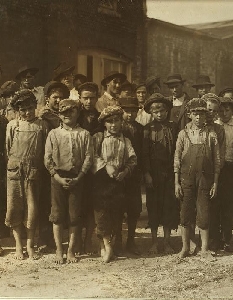}
\caption*{Hine}
\end{subfigure}
\hspace{-.3em}
\begin{subfigure}{.13\textwidth}
\centering
\includegraphics[width=1\columnwidth]{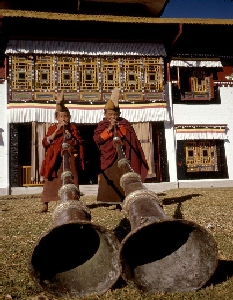}
\caption*{Kandell}
\end{subfigure}
\hspace{-.3em}
\begin{subfigure}{.13\textwidth}
\centering
\includegraphics[width=1\columnwidth]{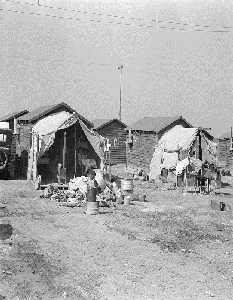}
\caption*{Lange}
\end{subfigure}
\hspace{-.3em}
\begin{subfigure}{.13\textwidth}
\centering
\includegraphics[width=1\columnwidth]{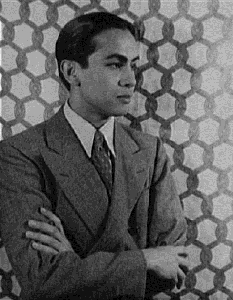}
\caption*{Van Vechten}
\end{subfigure}
\vspace{-0.6em}
\caption{Top ten objects and scenes for select photographers, and
  sample images.}
\label{table:topten}
\vspace{1.5em}
\end{table*}

\subsection{Schools of thought}
 
\ctt{Taking the idea of photographic style one step further, we wanted to see if meaningful genres or ``schools of thought'' of photographic style could be inferred from our results. 
We know that twelve of the photographers in our dataset were members of the Magnum Photos cooperative. We cluster the H-Pool5 features for all 41 photographers into a dendrogram, using agglomerative clustering, and discover that \akk{nine} of those twelve cluster together tightly, with only one non-Magnum photographer in their cluster. We find that \akk{three of the four} founders of Magnum form their own even tighter cluster. Further, five photographers in our dataset that were employed by the FSA are grouped in our dendrogram, and two portrait photographers (Van Vechten and Curtis) appear in their own cluster. See the supplementary file for the figure. These results indicate that our techniques are not only useful for describing individual photographers but can also be used to situate photographers in broader ``schools of thought.''}

\subsection{New photograph generation}
\label{sec:gen}

Our experimental results demonstrated that object and scene information is useful for distinguishing between photographers. Based on these results, we wanted to see whether we could take our photographer models yet another step further by generating new photographs imitating photographers' styles. Our goal was to create ``pastiches'' assembled by cropping objects out of each photographer's data and pasting them in new scenes obtained from Flickr. We first learned a probability distribution over the 205-scene types detected by Hybrid-CNN for each photographer. We then learned a distribution of objects and their most likely spatial location for each photographer, conditioned on the scene type. To do this, we trained a Fast-RCNN \cite{girshick2015fast} object detector on 25 object categories which frequently occurred across all photographers in our dataset using data we obtained from ImageNet. We then sampled from our joint probability distributions to choose which scene to use and which objects should appear in it and where. We \akkk{randomly selected} a detection (in that photographer's data) for each object probabilistically selected to appear, then cropped out the detection and segmented the cropped region using \cite{li2014secrets}. We inserted \akkk{the segment into the pastiche} according to that photographer's spatial model for that object.

\begin{figure*}[t]
\vspace{-1em}
\begin{center}
\begin{subfigure}{.16\linewidth}
\centering
\includegraphics[width=1\linewidth]{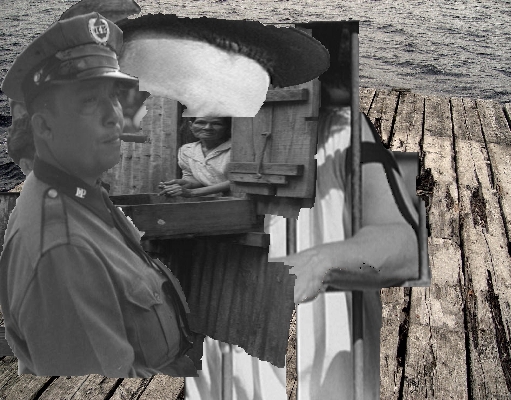}
\caption{Delano}
\label{fig:generated_images:a}
\end{subfigure}
\begin{subfigure}{.16\linewidth}
\centering
\includegraphics[width=1\linewidth]{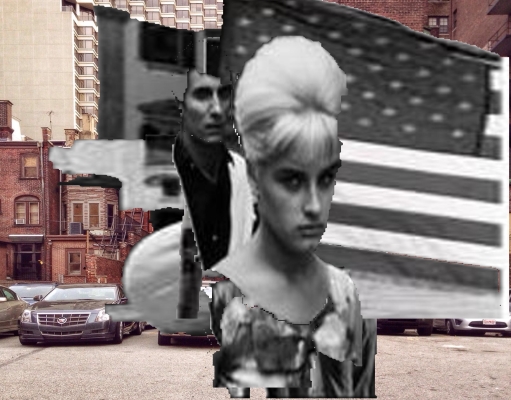}
\caption{Erwitt}
\label{fig:generated_images:b}
\end{subfigure}
\begin{subfigure}{.16\linewidth}
\centering
\includegraphics[width=1\linewidth]{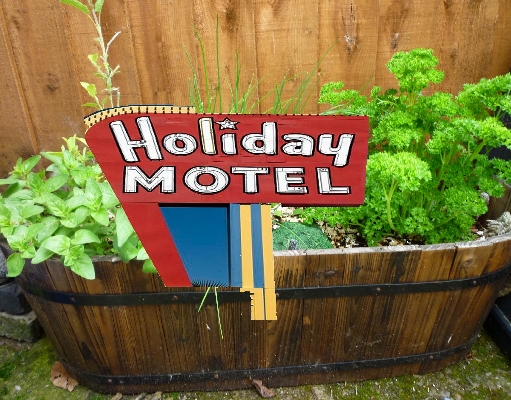}
\caption{Highsmith}
\label{fig:generated_images:c}
\end{subfigure}
\begin{subfigure}{.16\linewidth}
\centering
\includegraphics[width=1\linewidth]{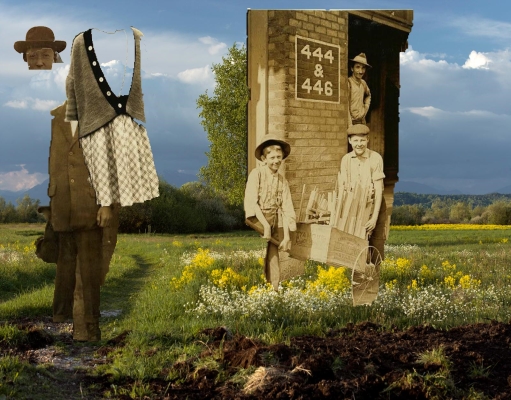}
\caption{Hine}
\label{fig:generated_images:d}
\end{subfigure}
\begin{subfigure}{.16\linewidth}
\centering
\includegraphics[width=1\linewidth]{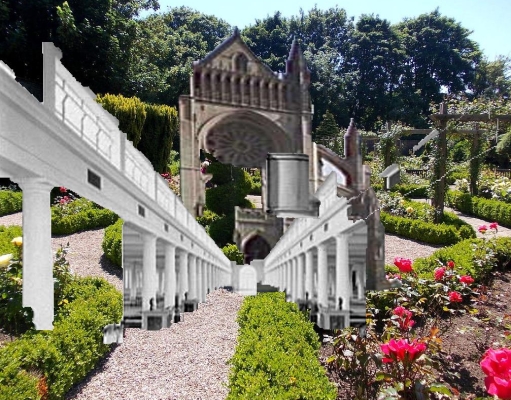}
\caption{Horydczak}
\label{fig:generated_images:e}
\end{subfigure}
\begin{subfigure}{.16\linewidth}
\centering
\includegraphics[width=1\linewidth]{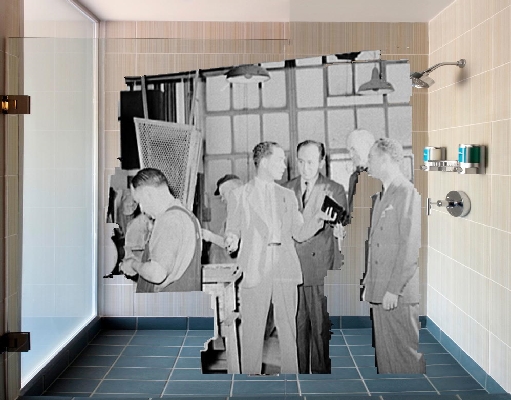}
\caption{Rothstein}
\label{fig:generated_images:f}
\end{subfigure}
\end{center}

\vspace{-2em}

\begin{center}
\begin{subfigure}{.08\linewidth}
\centering
\includegraphics[width=1\linewidth]{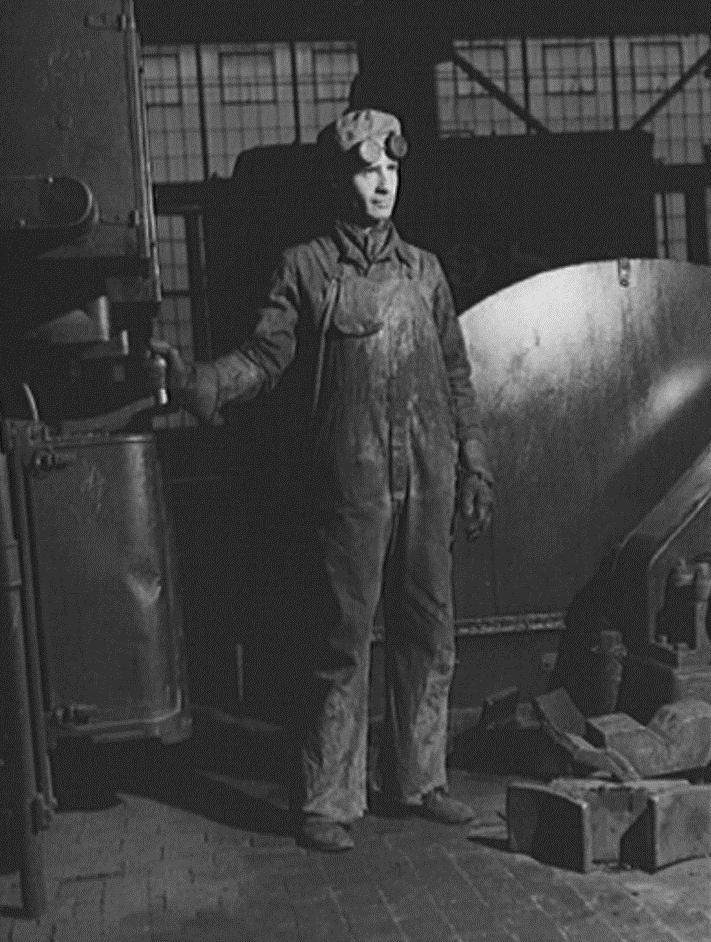}
\end{subfigure}
\begin{subfigure}{.08\linewidth}
\centering
\includegraphics[width=1\linewidth]{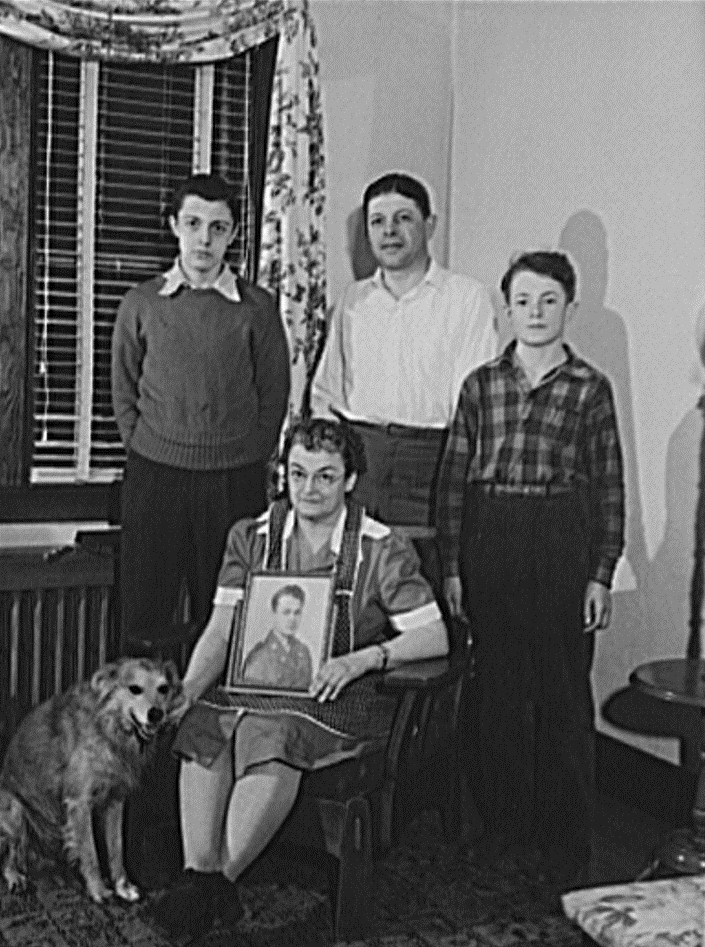}
\end{subfigure}
\begin{subfigure}{.16\linewidth}
\centering
\includegraphics[width=1\linewidth]{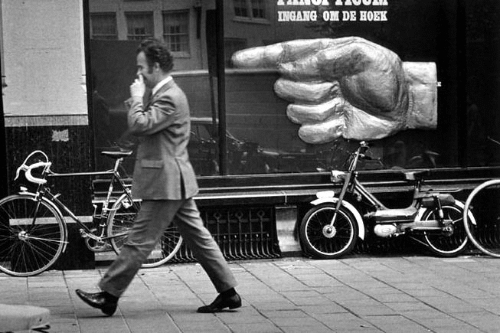}
\end{subfigure}
\begin{subfigure}{.16\linewidth}
\centering
\includegraphics[width=1\linewidth]{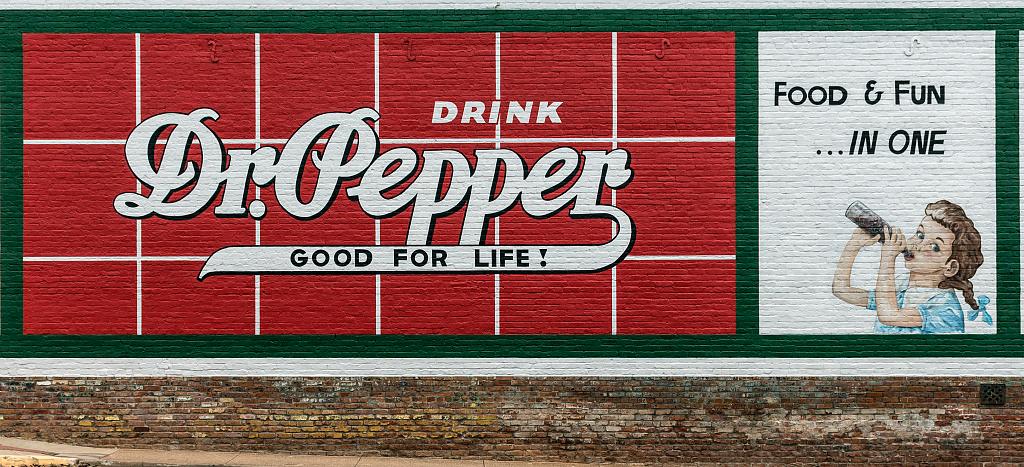}
\end{subfigure}
\begin{subfigure}{.16\linewidth}
\centering
\includegraphics[width=1\linewidth]{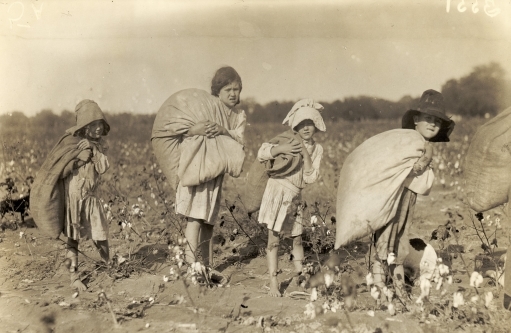}
\end{subfigure}
\begin{subfigure}{.08\linewidth}
\centering
\includegraphics[width=1\linewidth]{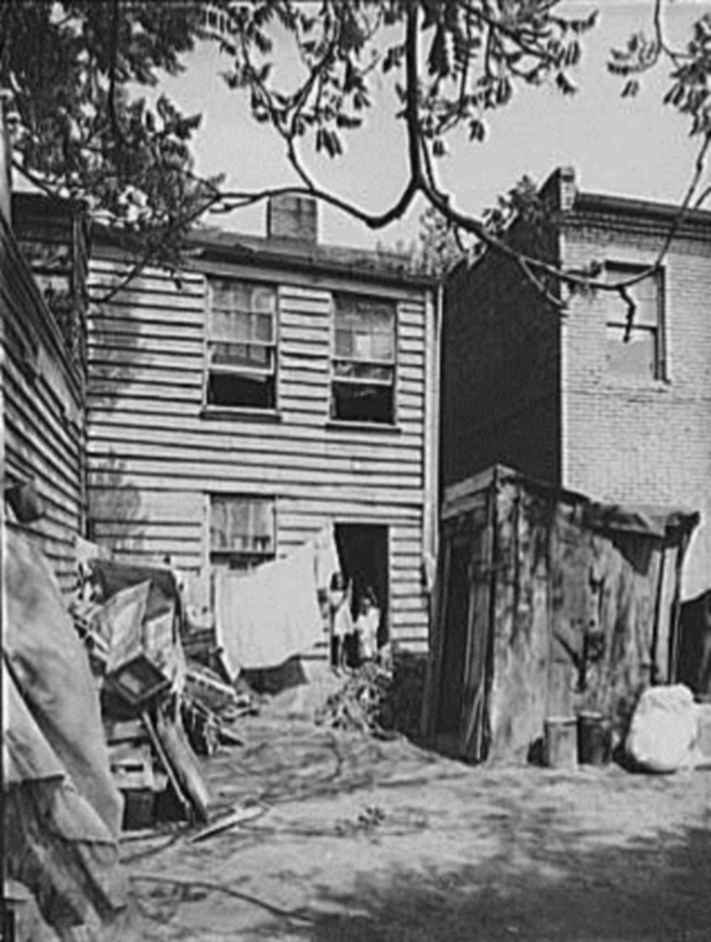}
\end{subfigure}
\begin{subfigure}{.08\linewidth}
\centering
\includegraphics[width=1\linewidth]{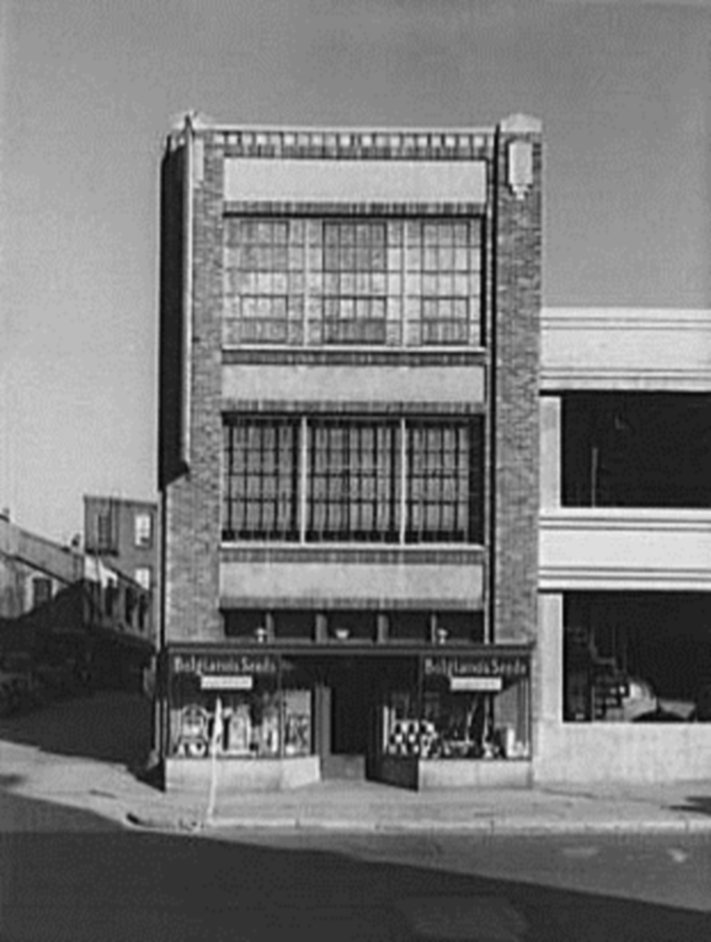}
\end{subfigure}
\begin{subfigure}{.16\linewidth}
\centering
\includegraphics[width=1\linewidth]{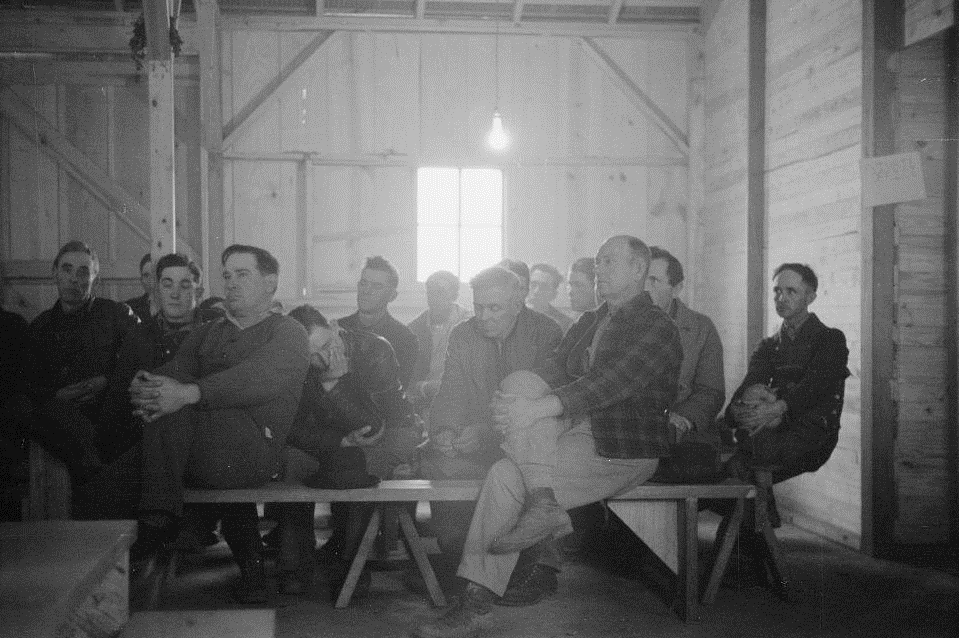}
\end{subfigure}

\end{center}
\vspace{-1em}
\caption{Generated images for six photographers (top row) and real photographs by these authors (bottom row). Although results are preliminary,  we observe interesting similarities between the synthetic and real work.}
\label{fig:generated_images}
\vspace{-1em}
\end{figure*}

We show \akkk{six pastiches generated using this approach} in Fig.\ \ref{fig:generated_images}. The top row shows generated images for six photographers, and the bottom shows real images from the corresponding photographer that resemble the generated ones.
For example, Delano takes portraits of individuals in uniforms and of ``common people,'' Erwitt photographs people in street scenes without their knowledge or participation, and Rothstein photographs people congregating. Highsmith captures large banner ads and Americana, Hine children working in poor conditions, and Horydczak buildings and architecture.
While these are preliminary results, we see similarities between the synthetic and authentic photos.

\section{Conclusion} \label{conclusion}
In this paper, we have proposed the novel problem of photograph authorship attribution. To facilitate research on this problem, we created a large dataset of \ctt{181,948} images by renowned photographers. In addition to tagging each photo with the photographer, the dataset also provides rich metadata which could be useful for future research in computer vision on a variety of tasks. 

Our experiments reveal that high-level features perform significantly better overall than low-level features \ctt{or humans.} \ctt{While our trained CNN, PhotographerNET, performs reasonably well, early proto-object and scene-detection features perform significantly better.} The inclusion of scene information provides moderate gains over the purely object-driven approach explored by \cite{ris, lsc}. 
We also provide an approach for performing qualitative analysis on the photographers by determining which objects respond strongly to each photographer in the feature values and learned classifier weights. Using these techniques, we were able to draw interesting conclusions about the photographers we studied \ctt{as well as broader ``schools of thought.''}
\akkk{We also showed initial results for a method that creates new photographs in the spirit of a given author.}

\akkk{In the future, we will develop further applications of our approach, \eg teaching humans to better distinguish between the photographers' styles.}
We will also continue our work on using our models to generate novel photographs of known photographers' styles.

\vspace{-0.3cm}
\paragraph{Acknowledgement.} \akkk{This work used the Extreme Science and Engineering Discovery Environment (XSEDE) 
and the Data Exacell at the Pittsburgh Supercomputing Center (PSC), 
supported by National Science Foundation grants ACI-1053575 and ACI-1261721.}


{\fontsize{7.5}{9.5}\selectfont 
\bibliographystyle{ieee}
\bibliography{photo}
}

\end{document}